\def\year{2018}
\relax
\documentclass[letterpaper]{article}
\usepackage{aaai18}
\usepackage{amsfonts}       
\usepackage{nicefrac}       
\usepackage{microtype}      

\usepackage{hyperref}
\usepackage{url}
\usepackage[table]{xcolor}
\usepackage{bbm}
\usepackage{booktabs}
\usepackage[T1]{fontenc}
\usepackage{fix-cm}
\usepackage{array}
\usepackage{epsfig}
\usepackage{dsfont}
\usepackage{multirow}

\usepackage{times}
\usepackage{helvet}
\usepackage{courier}
\usepackage{graphicx}
\usepackage{bm}
\usepackage{amsmath,amssymb} 
\usepackage{color}
\usepackage{bbm}
\usepackage{epstopdf}
\usepackage{caption}
\usepackage{subcaption}
\usepackage{enumitem}
\usepackage{calc}
\usepackage{multirow}
\usepackage{xspace}
\usepackage{booktabs}
\usepackage{mathrsfs}
\usepackage{array}

\newcommand{\figref}[1]{Fig\onedot~\ref{#1}}
\newcommand{\equref}[1]{Eq\onedot~\eqref{#1}}
\newcommand{\secref}[1]{Sec\onedot~\ref{#1}}

\newcommand{\ve}[1]{{\mathbf #1}} 
\newcommand{\hua}[1]{{\mathcal #1}}
\newcommand{\scr}[1]{{\mathcal #1}}

\DeclareRobustCommand\onedot{\futurelet\@let@token\@onedot}
\def\onedot{\ifx\@let@token.\else.\null\fi\xspace}
\def\eg{\emph{e.g.}} 

\def\any{\forall}
\def\ie{\emph{i.e.}}

\def\etc{\emph{etc}\onedot} 

\def\wrt{w.r.t\onedot}

\makeatletter
\newcommand{\thickhline}{%
    \noalign {\ifnum 0=`}\fi \hrule height 1pt
    \futurelet \reserved@a \@xhline
}
\newcolumntype{"}{@{\hskip\tabcolsep\vrule width 1pt\hskip\tabcolsep}}
\makeatother

\begin{document}
\title{Unsupervised Learning of Geometry from Videos with Edge-aware Depth-Normal Consistency}

\author{
Zhenheng Yang$^{1}$~~Peng Wang$^{2}$~~Wei Xu$^{2}$~~Liang Zhao$^{2}$~~Ramakant Nevatia$^{1}$\\
\\
\texttt{zhenheny@usc.edu~~\{wangpeng54,wei.xu,zhaoliang07\}@baidu.com~~nevatia@usc.edu}\\
$^{1}$University of Southern California~~~~$^{2}$Baidu Research\\
}


\copyrighttext{}
\maketitle

\begin{abstract}
    Learning to reconstruct depths in a single image by watching unlabeled videos via deep convolutional network (DCN) is attracting significant attention in recent years, \eg \cite{zhou2017unsupervised}. 
    In this paper, we introduce a surface normal representation for unsupervised depth estimation framework. Our estimated depths are constrained to be compatible with predicted normals, yielding more robust geometry results. 
    Specifically, we formulate an edge-aware depth-normal consistency term, and solve it by constructing a depth-to-normal layer and a normal-to-depth layer inside of the DCN. 
    The depth-to-normal layer takes estimated depths as input, and computes normal directions using cross production based on neighboring pixels. Then given the estimated normals, the normal-to-depth layer outputs a regularized depth map through local planar smoothness. Both layers are computed with awareness of edges inside the image to help address the issue of depth/normal discontinuity and preserve sharp edges.
    Finally, to train the network, we apply the photometric error and gradient smoothness for both depth and normal predictions.
    We conducted experiments on both outdoor (KITTI) and indoor (NYUv2) datasets, and show that our algorithm vastly outperforms state of the art, which demonstrates the benefits from our approach.
\end{abstract}

\vspace{-0\baselineskip}
\section{Introduction}
\label{sec:intro}
\vspace{-0\baselineskip}
\footnote{The work is done during Zhengheng Yang's internship at Baidu}
Human beings are highly competent in recovering the 3D geometry of observed natural scenes at a very detailed level in real-time, even from a single image. 
Being able to do reconstruction for monocular images can be widely applied to large amount of real applications such as augmented reality and robotics.


One group of approaches solve this problem by feature matching and estimating camera and scene geometries, \eg structure from motion (SFM) \cite{wu2011visualsfm} \etc, or color matching, \eg DTAM \cite{NewcombeLD11}. But these techniques are sensitive to correct matching and are ineffective in homogeneous areas. 
\begin{figure}
\centering
\includegraphics[width=0.48\textwidth, height=0.18\textwidth]{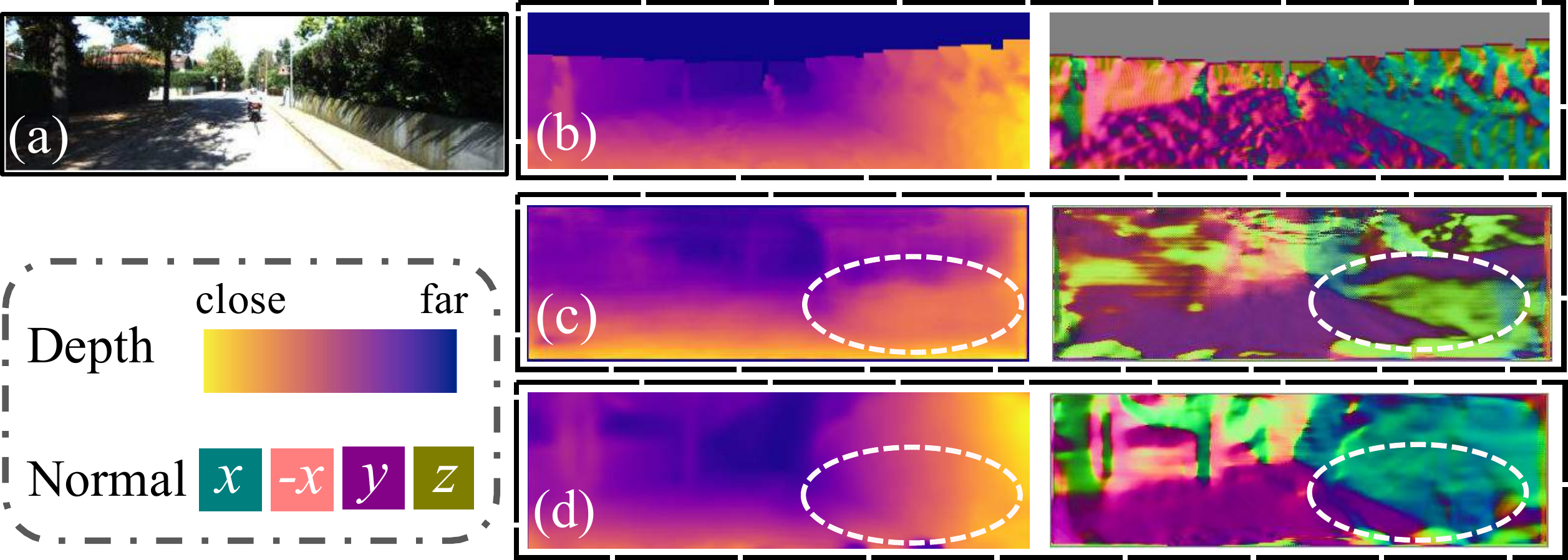}
\caption{Comparison between \protect\cite{zhou2017unsupervised} and our results with depth-normal consistency. Top to bottom: (b) Ground truth depths (left) and normals (right). (c) Results from \protect\cite{zhou2017unsupervised}. (d) Our results. In the circled region, \protect\cite{zhou2017unsupervised} fails to predict scene structure as shown by estimated normals, while ours correctly predict both depths and normals with such consistency.}
\vspace{-1.0\baselineskip}
\label{fig:visual_comparison}
\end{figure}
Another way to do 3D reconstruction is by a learning based method, where the reconstruction cues can be incrementally discovered by learning from videos. Currently, with the development of pixel-wise prediction such as fully convolutional network (FCN) \cite{long2015fully}, supervised learning of depth, \eg \cite{eigen2014depth}, achieved impressive results over public datasets like KITTI \cite{geiger2012we}, NYUv2 \cite{silberman2012indoor} and SUN3D \cite{xiao2013sun3d}. 
Nevertheless, collecting ground truth depth is almost impossible for random videos.  It is hard for the supervisedly learned models to generalize on videos of different scenes.

We can, instead, try to solve this problem in an unsupervised way by imposing 3D scene geometric consistency between video frames. There has been works in this line, \cite{zhou2017unsupervised} propose a single image depth FCN learning from videos. In their training, rather than using ground truth depth, they warp the target image to other consecutive video frames based on the predicted depths and relative motions, and match the photometry between the warped frames and observed frames (detailed in \secref{sec:preliminaries}). Then, the matching errors are used as the supervision of the depth prediction. Similar idea was applied in depth prediction when stereo pairs are available~\cite{GargBR16,godard2016unsupervised}.

Altough those methods are able to do single image depth estimation, the results do not well represent the scene structure, especially when visulized with computed normals, as shown in \figref{fig:visual_comparison}(c).
This is mostly due to that photometric matching is ambiguous, \ie a pixel in source frames can be matched to multiple similar pixels in target frames. Although researchers usually apply smoothness of depths \cite{zhou2017unsupervised} to reduce the ambiguity, it is often a weak constraint over neighboring pixels, which potentially have similar colors, thus yielding inconsistent normal results.

Our work falls in the scope of learning based 3D reconstruction of a single image trained on monocular videos, following the work of \cite{zhou2017unsupervised}. But we have a step further towards learning a regularized 3D geometry with explicit awareness of normal representation.
We are motivated by the fact that human beings are more sensitive in normal directions compared to depth estimation. For instance, one could precisely point out the normal direction of surface at each pixel of a single image while could only roughly know the absolute depth. 

Thus, we incorporate an edge-aware depth-normal consistency constraint inside the network which better regularizes the learning of depths (\secref{sec:approach}). 
There are several advantages of having normal estimated. For instance, it gives explicit understanding of normal for learned models.  In addition, it provides higher order interaction between estimated depths, which is beyond local neighbor relationships. Last, additional operations, \eg Manhattan assumption, over normals could be further integrated. As depth/normal discontinuity often appear at object edges in the image, we incoporate the image edges in this constraint to compensate.
As shown at \figref{fig:visual_comparison}(d), with such a constraint, our recovered geometry is comparably better. We did extensive experiments over the public KITTI and NYUv2 datasets, and show our algorithm can achieve relative 20$\%$ improvement over the state-of-the-art method on depth estimation and 10$\%$ improvement on predicted normals. More importantly, the training converges around 3$\times$ faster. These demonstrate the efficiency and effectiveness of our approach.

\vspace{-0\baselineskip}
\section{Related Work}
\label{sec:related}

\textbf{~~~Structure from motion and single view geometry.}
As discussed in \secref{sec:intro}, geometry based methods, such as SFM~\cite{wu2011visualsfm}, ORB-SLAM~\cite{mur2015orb}, DTAM~\cite{NewcombeLD11}, rely on feature matching, which could be effective and efficient in many cases. 
However, they can fail at low texture, or drastic change of visual perspective \etc. More importantly, it can not extend to single view reconstruction where humans are good at.
Traditionally, specific rules are developed for single view geometry. Methods are dependent on either computing vanishing point~\cite{HoiemEH07}, following rules of BRDF~\cite{prados2006shape}, or abstract the scenes with major plane and box representations~\cite{DBLP:conf/iccv/SchwingFPU13,DBLP:conf/3dim/SrajerSPP14} \etc. Those methods can only obtain sparse geometry representations, and some of them require certain assumptions (\textit{e.g.} Lambertian, Manhattan world).

\textbf{Supervised single view geometry via CNN.}
With the advance of deep neural networks and their strong feature representation, dense geometry, i.e., pixel-wise depth and normal maps, can be readily estimated from a single image~\cite{wang2015designing,eigen2015predicting,laina2016deeper}. The learned CNN model shows significant improvement comparing to other strategies based on hand-crafted features~\cite{karsch2014depth,ladicky2014pulling,zeisl2014discriminatively}. Others tried to improve the estimation further by appending a conditional random field (CRF)~\cite{DBLP:conf/cvpr/WangSLCPY15,Liu_2015_CVPR,li2015depth}. 
However, most works regard depth and normal predictions as independent tasks. \cite{peng2016depth} point out their correlations over large planar regions, and regularize the prediction using a dense CRF~\cite{kong2015intrinsic}, which improved the results on both depth and normal. However, all those methods require densely labeled ground truths, which are expensive to label in natural environments.

\textbf{Unsupervised single view geometry.}
Videos are easy to obtain at the present age, while hold much richer 3D information than single images. Thus, it attracts lots of interests if single view geometry can be learned through feature matching from videos. Recently, several deep learning methods have been proposed based on such an intuition. Deep3D \cite{xie2016deep3d} learns to generate the right view from the given left view by supervision of a stereo pair. In order to do back-propagation to depth values, it quantizes the depth space and trains to select the right one. 
Concurrently, \cite{GargBR16} applied the similar supervision from stereo pairs, while the depth is kept continuous, They apply Taylor expansion to approximate the gradient for depth. \cite{godard2016unsupervised} extend Garg's work by including depth smoothness loss and left-right depth consistency. Most recently, \cite{zhou2017unsupervised} induces camera pose estimation into the training pipeline, which makes depth learning possible from monocular videos. And they come up with an explainity mask to relieve the problem of moving object in rigid scenes.
At the same time, \cite{kuznietsov2017semi} proposed a network to include modeling rigid object motion. Although vastly developed for depth estimation from video, normal information, which is also highly interesting for geometry prediction, has not been considered inside the pipeline. This paper fills in the missing part, and show that normal can serve as a natural regularization for depth estimation, which significantly improves the state-of-the-art performance. Finally, with our designed loss, we are able to learn the indoor geometry where \cite{zhou2017unsupervised} usually fails to estimate.


\vspace{-0\baselineskip}
\section{Preliminaries}
\label{sec:preliminaries}
\vspace{-0\baselineskip}

\begin{figure*}[!htp]
\centering
\includegraphics[width=\textwidth]{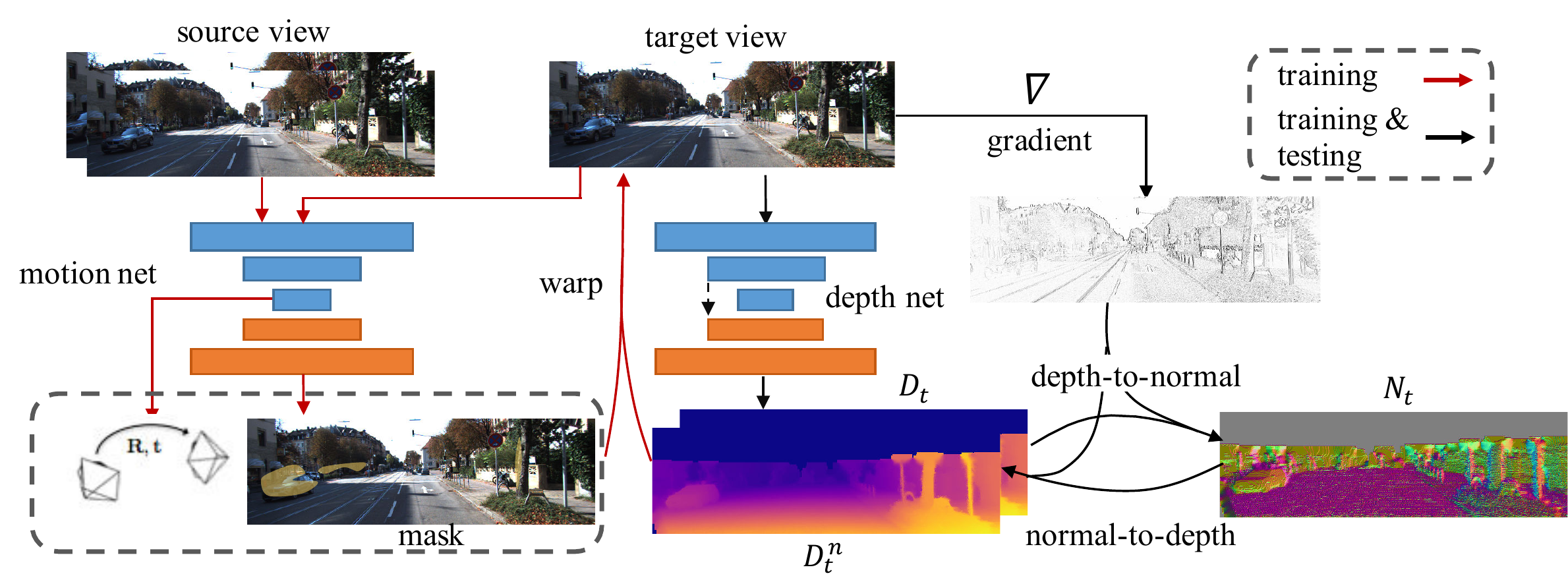}
\caption{Framework of our approach.}
\label{fig:pipeline}
\vspace{-.3\baselineskip}
\end{figure*}

\begin{figure}[!htp]
\centering
\includegraphics[width=0.5\textwidth]{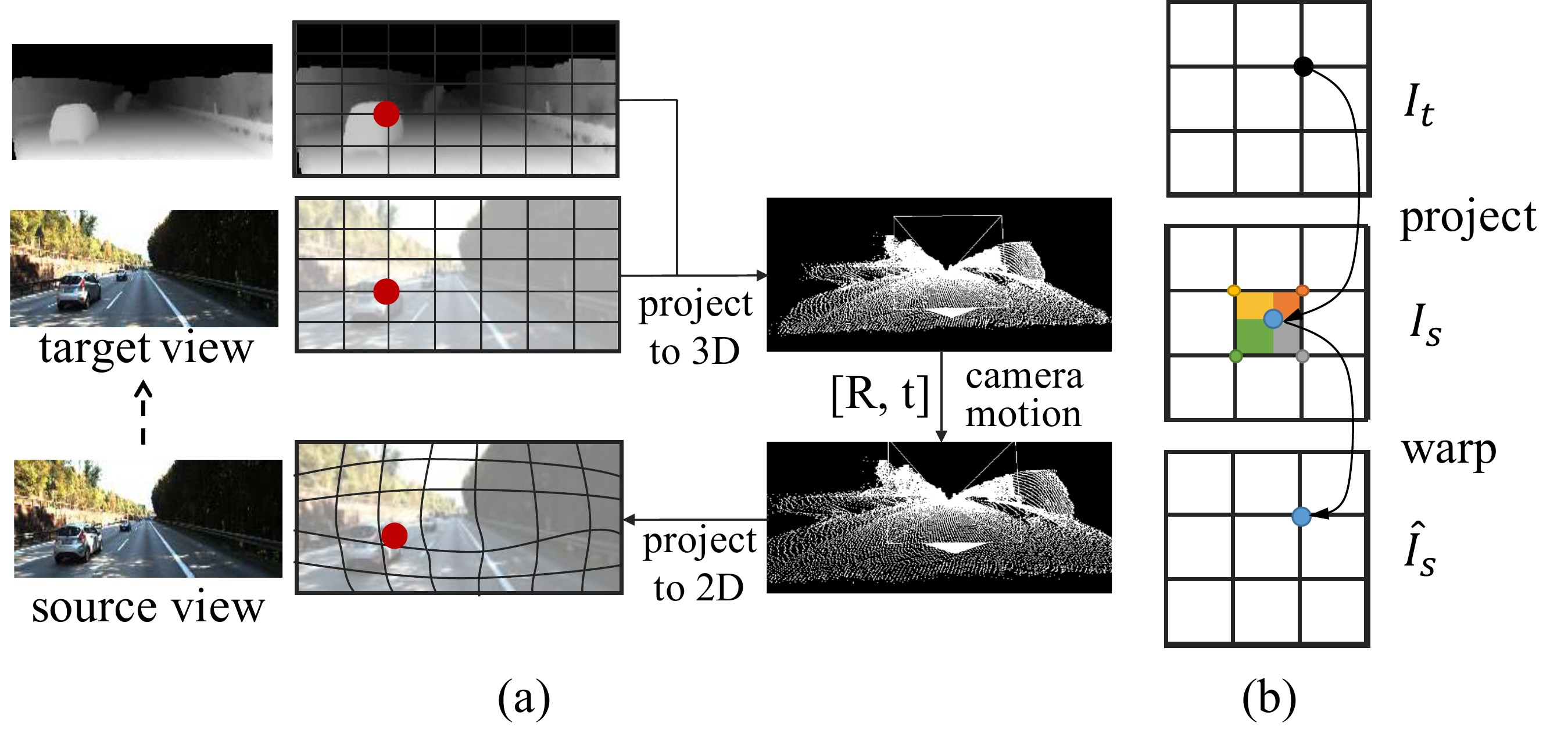}
\caption{Illustraion of (a) 3D inverse warping and (b) bilinear interpolation.}
\label{fig:3d_warping}
\vspace{-1.1\baselineskip}
\end{figure}

In order to make the paper self-contained, we first introduce several preliminaries proposed in the unsupervised learning pipelines \cite{zhou2017unsupervised,godard2016unsupervised}. The core idea behind, as discussed in Sec. \ref{sec:related}, is inverse warping from target view to source view with awareness of 3D geometry, as illustrated in Fig. \ref{fig:3d_warping}(a), which we will elaborate in the following paragraphs.

\textbf{Perspective projection between multiple views.}
Let $D(x_t)$ be the depth value of the target view at image coordinate $x_t$, and $\ve{K}$ be the intrinsic parameter of the camera. Suppose the relative pose from the target view to source view is a rigid transformation $\ve{T}_{t\rightarrow s} = [\ve{R} | \ve{t}] \in \hua{S}\hua{E}(3)$, and $h(x)$ is the homogeneous coordinate given $x$. The perspective warping to localize corresponding pixels can be formulated as, 
\begin{align}
\label{eqn:warp}
D(x_s)h(x_s) = \ve{K}\ve{T}_{t\rightarrow s}D(x_t)\ve{K}^{-1}h(x_t),
\end{align}
and the image coordinate $x_s$ can be obtained by dehomogenisation of $D(x_s)h(x_s)$. Thus, $x_s$ and $x_t$ is a pair of matching coordinates, and we are able to compare the similarity between the two to validate the correctness of structure.

\textbf{Photometric error from view synthesis.} 
\label{chap:warping}
Given pixel matching pairs between target and source views, \ie $I_t$ and $I_s$, we can synthesis a target view $\hat{I_s}$ from the given source view through bilinear interpolation~\cite{GargBR16}, as illustrated in Fig. \ref{fig:3d_warping}(b). 
Then, under the assumption of Lambertian and a static rigid scene, the average photometric error is often used to recover the depth map $D$ for the target view and the relative pose. 
However, as pointed out by \cite{zhou2017unsupervised}, this assumption is not always true, due to the fact of moving objects and occlusion. An explainability mask $\ve{M}$ is induced to compensate for this. Formally, the masked photometric error is,
\begin{align}
\label{eqn:photometric}
& \scr{L}_{vs}(D, \hua{T}, \hua{M}) = \sum_{s=1}^{S}\sum_{x_t}\ve{M}_s(x_t)|I_t(x_t) - \hat{I_s}(x_t)|, \nonumber \\
& \text{s.t. ~} \any x_t, s\text{~~} \ve{M}_s(x_t)\in [0, 1],\text{~} D(x_t) > 0
\end{align}
where $\{\hat{I_s}\}_{s=1}^{S}$ is the set of warped source views, and $\hua{T}$ is a set of transformation from target view to each of the source views. 
$\hua{M} = \{\ve{M}_s\}$ is a set of explainability masks, and $\ve{M}_s(x_t) \in [0, 1]$ weights the error at $x_t$ from source view $s$.



\textbf{Regularization.} 
As mentioned in \secref{sec:intro}, supervision based solely on photometric error is ambiguous. One pixel could match to multiple candidates, especially in low-texture regions. In addition, there is trivial solution for explainability mask by setting all values to zero. Thus, to reduce depth ambiguity and encourage non-zero masks, two regularization terms are applied, 

\begin{align}
\label{eqn:regular}
\scr{L}_{s}(D, 2) &= \sum_{x_t}\sum_{d \in {x, y}}|\nabla^{2}_dD(x_t)|e^{-\alpha|\nabla_dI(x_t)|} & \nonumber \\
\scr{L}_{m}(\hua{M}) &= -\sum_s\sum_{x_t}\log P(\ve{M_s}(x_t) = 1) &
\end{align}
$\scr{L}_{s}(D, 2)$ is a spatial smoothness term penalizes L1 norm of second-order gradients of depth along both x and y directions, encouraging depth values to align in planar surface when no image gradient appears. Here, the number $2$ represents the 2nd order for depth. $\scr{L}_{m}(\hua{M})$ is cross-entropy between the masks and maps with value 1.


Finally, a multi-scale strategy is applied to the depth output, and the total loss for depth estimation from videos is a joint functional from previous terms,
\begin{align}
\label{eqn:full}
\scr{L}_{o}(\{D_l\}, \hua{T}, \hua{M}) =& \sum_l\{\scr{L}_{vs}(D_l,\hua{T},\hua{M}) + \lambda_s\scr{L}_{s}(D_l) \nonumber\\
&\text{~~~~~} + \lambda_m\scr{L}_{m}(\hua{M}_l)\}
\end{align}

Given the objective functional, the photometric error can be back-propagated to depth, pose and mask networks by applying the spatial transform operation as proposed by~\cite{jaderberg2015spatial}, which supervises the learning process.

\vspace{-0\baselineskip}
\section{Geometry estimation with edge-aware depth-normal consistency}
\label{sec:approach}
\vspace{-0\baselineskip}

In our scenario, given a target image $I$, we aim at learning to estimate both depths and normals simultaneously. Formally, let $N$ be the predicted normals from our model, we embed it into the training pipeline and make it a regularization for depths estimation $D$, which helps to train a more robust model.

\vspace{-0\baselineskip}
\subsection{Framework}
\label{sub:framework}
\vspace{-0\baselineskip}

\figref{fig:pipeline} illustrates an overview of our approach. For training, we apply supervision from view synthesis following \cite{zhou2017unsupervised}. Specifically, the depth network (middle) takes only the target view as input, and
outputs a per-pixel depth map $D_t$, based on which a normal map $N_t$ is generated by the depth-to-normal layer. Then, given the $D_t$ and $N_t$, a new depth map $D_t^n$ is estimated from the normal-to-depth layer using local orthogonal compatibility between depth and normals. Both of the layers takes in image gradient to avoid non-compatible pixels involving in depth and normal conversion (detailed in \secref{sub:depth_and_normal_orthogonality}).
Then, the new depth map $D_t^n$, combined with poses and mask predicted from the motion network (left), are then used to inversely warp the source views to reconstruct the target view, and errors are back propagated through both networks. Here the normal representation naturally serves as a regularization for depth estimation. Finally, for training loss, additional to the usually used photometric reconstruction loss, we also add in smoothness over normals, which induces higher order interaction between pixels (\secref{sub:training_losses})

With the trained model, given a new image,  we infer per-pixel depth value and then compute the normal value, yielding consistent results between the two predictions.

\vspace{-0\baselineskip}
\subsection{Depth and normal orthogonality.}
\label{sub:depth_and_normal_orthogonality}
\vspace{-0\baselineskip}

In reconstruction, depth and normal are two strongly correlated information, which follows locally linear orthogonality. Formally, for each pixel $x_i$, such a correlation can be written as a quadratic minimization for a set of linear equations,
\begin{align}
\label{eq:orthognal}
&\scr{L}_{x_i}(D, N) = ||[\cdots,\omega_{ji}(\phi(x_j) - \phi(x_i)), \cdots]^T  N(x_i)||^2, \nonumber \\
&~\text{where~} \phi(x) = D(x)\ve{K}^{-1}h(x), \text{~} \|N(x_i)\|_2 = 1, \nonumber\\
&~\text{~~~~~~~~} \omega_{ji} > 0 \text{~~if~~} x_j \in \hua{N}(x_i)
\end{align}
where $\scr{N}(x_i)$ is a set of predefined neighborhood pixels of $x_i$, and $N(x_i)$ is a $3 \times 1$ vector. $\phi(x)$ is the back projected 3D point from 2D coordinate $x$. $\phi(x_j) - \phi(x_i)$ is a difference vector in 3D, and $\omega_{ji}$ is used to weight the equation for pixel $x_j$ \wrt $x_i$ which we will elaborate later.

As discussed in Sec. \ref{sec:related}, most previous works try to predict the two information independently without considering such a correlation, while only SURGE~\cite{peng2016depth} proposes to apply the consistency by a post CRF processing only over large planar regions. In our case, we enforce the consistency over the full image, and directly apply it to regularize the network to help the model learning. Specifically, to model their consistency, we developed two layers by solving \equref{eq:orthognal}, \ie a depth-to-normal layer and a normal-to-depth layer. 

\textbf{Infer normals from depths.} 
\label{chap:d2n}
Given a depth map $D$, for each point $x_i$, in order to get $N(x_i)$. From \equref{eq:orthognal}, we need to firstly define neighbors $\hua{N}(x_i)$ and weights $\omega_{ji}$, and then solve the set of linear equations. To deal with the first issue, we choose to use the 8-neighbor convention to compute normal directions, which considerably more robust than the 4-neighbor convention. 
However, it is not always good to equally weight all pixels due to depth discontinuity or dramatic normal changes may occur nearby. Thus, for computing $\omega_{ji}$, we weight more for neighboring pixels $x_j$ having similar color with $x_i$, while weight less otherwise. Formally, in our case, it is computed as $\omega_{ji} = \exp\{-\alpha|I(x_j) - I(x_i)|\}$ and $\alpha = 0.1$. 

\begin{figure}
\centering
\includegraphics[width=0.5\textwidth]{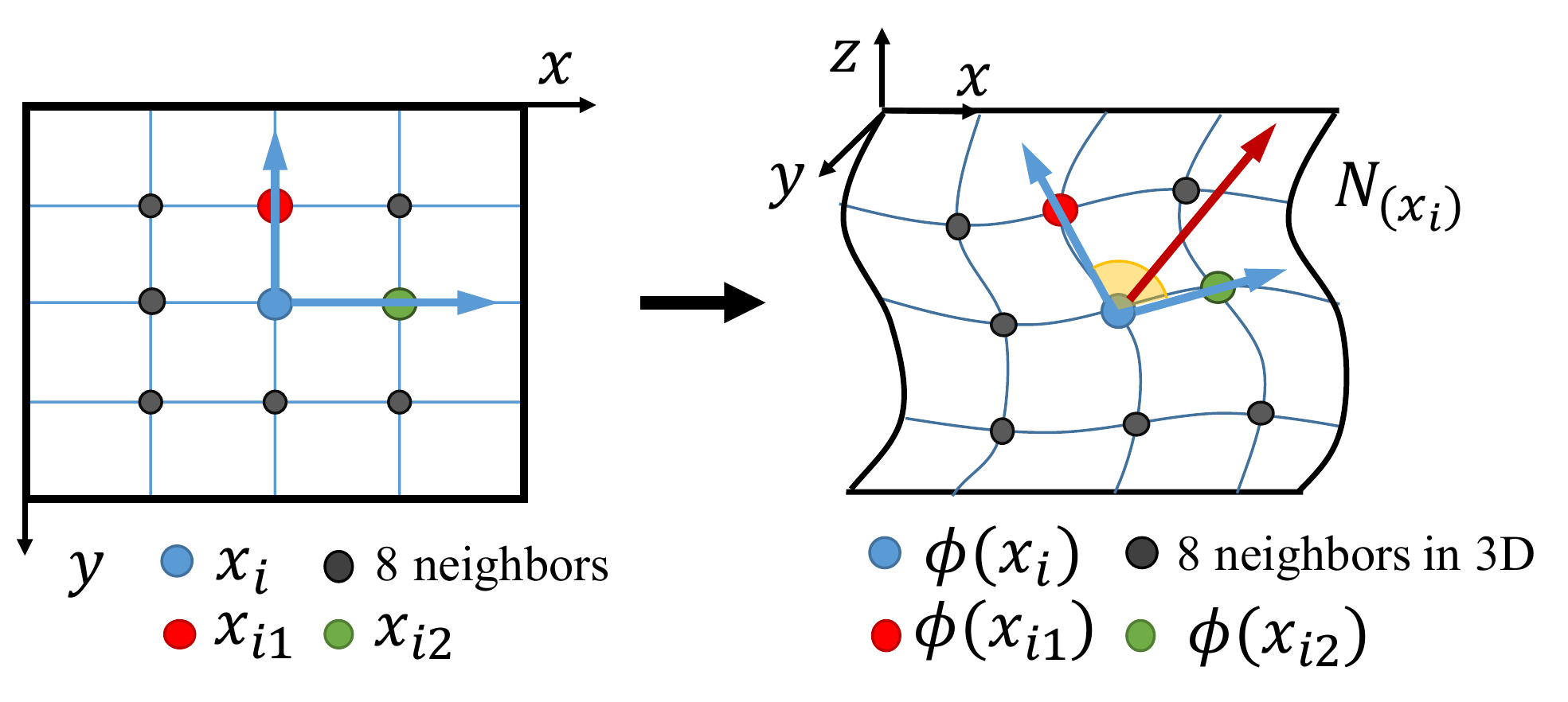}
\caption{Illustration of computing normal base on a pair of neighboring pixels. $x_i, x_{i1}, x_{i2}$ are 2D points, and 
$\phi(x_i), \phi(x_{i1}), \phi(x_{i2})$ are corresponding points projected to 3D space. 
The normal direction $N(x_i)$ is computed with cross product between $\phi(x_{i1}) - \phi(x_i)$ and $\phi(x_{i2}) - \phi(x_i)$.}
\vspace{-0.3\baselineskip}
\label{fig:d2n}
\end{figure}

For minimizing \equref{eq:orthognal}, one may apply a standard singular value decomposition (SVD) to obtain the solution. However, in our case, we need to embed such an operation in the network for training, and back-propagate the gradient respect to input depths. SVD is computationally non-efficient for back-propagation. Thus, we choose to use mean cross-product to approximate the minimization~\cite{jia2006using}, which is simpler and more efficient. 
Specifically, from the 8 neighbor pixels around $x_i = [m, n]$, we split them to 4 pairs, where each pair of pixels is perpendicular at 2D coordinate \wrt $x_i$, and in a counter clock-wise order, \ie $\hua{P}(x_i) = \{([m-1, n], [m, n+1]), \cdot, ([m+1, n-1], [m-1, n-1])\}$. 
Then, for each pair, cross product of their difference vector \wrt $x_i$ is computed, and the mean direction of the computed vectors is set as the normal direction of $x_i$. Formally, the solver for normals is written as, 
\begin{align}
\label{eq:cross}
&\ve{n} = \sum_{p\in\hua{P}}(\omega_{p_{0}, x_i}(\phi(p_{0}) - \phi(x_i)) \times \omega_{p_{1}, x_i}(\phi(p_{1}) - \phi(x_i))), \nonumber \\
&N(x_i) = \ve{n} / \|\ve{n}\|_2
\end{align}
The process of calculating the normal direction for $x_i$ using one pair of pixels is in Fig. \ref{fig:d2n}. 


\textbf{Compute depths from normals.} 
Due to the fact that we do not have ground truth normals for supervision, it is necessary to recover depths from normals to receive the supervision from photometric error as discussed in Sec.~\ref{sec:preliminaries}.
To recover depths, given normal map $N$, we still need to solve \equref{eq:orthognal}. However, there is no unique solution. Thus, to make it solvable, we provide an initial depth map $D_o$ as input, which might lack normal smoothness, \eg depth map from network output. Then, given $D_o(x_i)$, the depth solution for each neighboring pixel of $x_i$ is unique and can be easily computed. Formally, let $D_e(x_j | x_i) = \psi(D_o(x_i), N(x_i))$ be the solved depth value calculated for a neighbor pixel $x_j$ \wrt $x_i$. 
However, when computing over the full image, we still need to solve 8 equations jointly for each pixel of the 8 neighbors. Finally, by minimum square estimation (MSE), the solution for depth of $x_i$ is,
\begin{align}
\label{eq:depth}
D_n(x_j) = \sum_{i\in\hua{N}}\hat{\omega}_{ij}D_e(x_j | x_i), \text{~~}
\hat{\omega}_{ij} = \omega_{ij} / \sum_i{\omega_{ij}}
\end{align}

\vspace{-0\baselineskip}
\subsection{Training losses}
\label{sub:training_losses}
\vspace{-0.\baselineskip}

Given the consistency, in this section, we describe our training strategy. In order to supervise both the depth and normal predictions, we can directly apply the loss in \equref{eqn:full} by replacing the output from depth network $D_o$ with the output after our normal-to-depth layer $D_n$ to train the model. We show in our experiments (\secref{sec:experiments}), by doing this, we already outperform the previous state-of-the-art by around 10$\%$ in depth estimation using the same network architecture.

Additionally, with normal representation, we apply smoothness over neighboring normal values, which provides higher order interactive between pixels. Formally, the smoothness for normal has the same form as $\scr{L}_{s}$ in \equref{eqn:regular} for depth, while the first order gradient is applied, \ie $~\scr{L}_{s}(N, 1)$. 

Last but not the least, matching corresponding pixels between frames is another central factor to find correct geometry. Additional to the photometric error from matching pixel colors, matching image gradient is more robust to lighting variations, which was frequently applied in computing optical flow~\cite{li2017pyramidal}. 
In our case, we compute a gradient map of the target image and synthesized target images, and include the gradient matching error to our loss function. Formally, the loss is represented as,
\begin{equation}
\label{equ:gradient}
\scr{L}_{g}(D_n, \hua{T}, \hua{M}) = \sum_{s=1}^{S}\sum_{x_t}\ve{M}_s(x_t)\|\nabla I_t(x_t) - \nabla \hat{I_s}(x_t)\|_1, \nonumber
\end{equation}

In summary, our final learning objective for multi-scale learning is,
\begin{align}
\label{eq:full_loss}
&\scr{L}(\hua{D}, \hua{N}, \hua{T}, \hua{M}) = \scr{L}_{o}(\{D_{nl}\}, \hua{T}, \hua{M}) + \nonumber\\
&\text{~~~~~~~~~}\sum_l\{\lambda_g\scr{L}_{g}(D_{nl},\hua{T},\hua{M}) + \lambda_n\scr{L}_{s}(N_l, 1)\}
\end{align}
where $\hua{D} = \{D_{nl}\}$ and $\hua{N} = \{N_{l}\}$ are the set of depth maps and normal maps for the target view.


\textbf{Model training.} For network architecture, similar to \cite{zhou2017unsupervised} and \cite{godard2016unsupervised}, we adopt the DispNet \cite{mayer2016large} architecture with skip connections as in \cite{zhou2017unsupervised}. All \textit{conv} layers are followed by a ReLU activation except for the top prediction layer. We train the network from scratch; since too many losses at beginning could be hard to optimize, we choose a two stage training strategy by first train the network with $\scr{L}_{o}$ with 5 epochs and then fine-tune it with the full loss for 1 epoch. We provide ablation study of each term in our experiments.



 



\vspace{-0\baselineskip}
\section{Experiments}
\label{sec:experiments}
\vspace{-0\baselineskip}

In this section, we introduce implementation details, datasets, evaluation metrics. An ablation study of how much each component of the framework contributes and a performance comparison with other supervised or unsupervised methods are also presented.

\vspace{-0\baselineskip}
\subsection{Implementation details.}
\vspace{-0\baselineskip}
Our framework is implemented with publicly available TensorfFlow \cite{abadi2016tensorflow} platform and has 34 million trainable variables in total. During training, Adam optimizer is applied with parameters $\beta_1 = 0.9$, $\beta_2=0.000$, $\epsilon=10^{-8}$. Learning rate and batch size are set to be $2\times10^{-3}$ and $4$ respectively. Batch normalization \cite{ioffe2015batch} is not used as we didn't observe a performance improvement with it. Following \cite{zhou2017unsupervised}, we use the same loss balancing for $\lambda_s, \lambda_m$, and correct the depth by a scale factor. We set $\lambda_n=1$ and $\lambda_g=\lambda_s$. 

The length of input sequence is fixed to be 3 and the input frames are resized to $128 \times 416$. The middle frame is treated as the target image and the other two are source images. Our network starts to show meaningful results after 3 epochs, and converges at the end of the 5th epoch. With a Nvidia Titan X (Pascal), the training process takes around 6 hours. The number of epochs and absolute time needed is much less than \cite{godard2016unsupervised} (50 epochs, 25 hours) and \cite{zhou2017unsupervised} (15 epochs).

\begin{table*}[t]
\centering
\caption{Depth performance of our framework variants on the KITTI split.}
\label{tbl:ablation}
\fontsize{8}{9}\selectfont
\bgroup
\def\arraystretch{1.2}
\begin{tabular}{c|c|c|c|c|c|c|c}
\thickhline
\multirow{2}{*}{Methods}  & \multicolumn{4}{c|}{Lower the better} & \multicolumn{3}{c}{Higher the better}                  \\ \cline{2-8} 
                          & Abs Rel  & Sq Rel  & RMSE  & RMSE log & $\delta < 1.25$ & $\delta < 1.25^2$ & $\delta < 1.25^3$ \\ \hline
Ours (no d-n)             & 0.208    & 2.286   & 7.462 & 0.297    & 0.693           & 0.875             & 0.948             \\
Ours (smooth no gradient) & 0.189    & 1.627   & 7.017 & 0.280    & 0.713           & 0.891             & 0.957             \\
Ours (no img grad for d-n)    & 0.179    & 1.566   & 7.247 & 0.272    & 0.720           & 0.895             & 0.959             \\
Ours (no normal smooth)   & 0.172    & 1.559   & 6.794 & 0.252    & 0.744           & 0.910             & 0.969             \\ \hline
\end{tabular}
\egroup
\vspace{-0\baselineskip}
\end{table*}

\vspace{-0.\baselineskip}
\subsection{Datasets and metrics}
\vspace{-0\baselineskip}
\textbf{~~~Training.}
Theorectically, our framework can be trained on any frame sequences captured with a monocular camera. To better compare with other methods, we evaluate on the popular KITTI 2015 \cite{geiger2012we} dataset. It is a large dataset suite for multiple tasks, including optical flow, 3D object detection, tracking, and road segmenations, \etc The raw data contains RGB and gray-scale videos, which are captured by stereo cameras from 61 scenes, with a typical image size of $1242 \times 375$.

In our experiments, videos captured by both left and right cameras are used for training, but treated independently. We follow the same training sequences as \cite{zhou2017unsupervised,eigen2014depth}, excluding frames from test scenes and static sequences. This results in 40,109 trainig sequences and 4431 validation sequences. Different from \cite{godard2016unsupervised}, no data augmentation has been performed.

\textbf{Testing.}
There are two sets of KITTI 2015 test data: (1) Eigen split contains 697 test images proposed by \cite{eigen2014depth}; (2) KITTI split contains 200 high-quality disparity images provided as part of official KITTI training set.  To better compare with other unsupervised and supervised methods, we present evaluations on both splits. 

The depth ground truth of Eigen split is generated by projecting 3D points scanned from Velodyne laser to the camera view. This produces depth values for less than 5\% of all pixels in the RGB images. To be consistent when comparing with other methods, the same crop as in \cite{eigen2014depth} is performed when testing. The depth ground truth of KITTI split contains sparse depth map with CAD models in place of moving cars. It provides better quality depth than projected Velodyne laser scanned points but has ambiguous depth value on object boundaries where the CAD model doesn't align with the images. The predicted depth is capped at 80 meters as in \cite{godard2016unsupervised} and \cite{zhou2017unsupervised}.

The normal ground truth for two splits is generated by applying our depth-to-normal layer on inpainted depth ground truth, where the same inpainting algorithm as \cite{silberman2012indoor} is used. For both depth and normal, following \cite{eigen2014depth}, only the pixels with laser ground truth are used.

\textbf{Metrics.} We apply the same depth evaluation and normal evaluation metrics as in \cite{eigen2015predicting}. For depth evaluation, we use the code provided by~\cite{zhou2017unsupervised} and for normal, we implement ourselves and verified the correctness through validating normal results of \cite{eigen2015predicting} over the NYUv2 dataset.









\vspace{-0\baselineskip}
\subsection{Ablation study}
\vspace{-0\baselineskip}
To investigate different components proposed in \secref{sec:approach}, we perform an ablation study by removing each one from our full model and evaluating on the KITTI split.

\textbf{Depth-normal consistency.} By removing normal-to-depth layer (\equref{eq:depth}), the inverse warping process (Sec. \ref{chap:warping}) takes an image and directly predicted depth map from the input. We show the performance at the row ``Ours (no d-n)" in Tab. \ref{tbl:ablation}. It is much worse than our full model on Kitti shown in Tab. \ref{tbl:sota}.
Notice that with depth-normal consistency, the network not only performs better but converges faster. In fact, our full model converges after 5 epochs, while the network without such consistency converges at 15th epoch.

\textbf{Image gradient in smoothness term.} To validate image gradient for depth and normal smoothness in \equref{eqn:regular}, 
we setting $\alpha=0$. The results is shown as ``Ours (smooth no gradient)" in Tab. \ref{tbl:ablation}. It makes less impoact than depth-normal consistency, but still helps the performance.

\textbf{Image gradient in normal-depth consistency.} We set $\omega=1$ in \equref{eq:orthognal}, thus there is no edge awareness in depth-normal consistency. As show at row ``Ours (no img grad for n-d)'', the results are again worse than our final results, which demonstrates the effectiveness by only enforcing the consistency between color similar pixels. 

\textbf{Normal smoothness.} Finally, by removing normal smoothness $\scr{L}_n$ in \equref{eq:full_loss}, we show the results at row ``Ours (no normal smooth)'' in Tab. \ref{tbl:ablation}, where it makes less impact for depth than other components, while still make reasonable contributions. However, it makes relatively more contributions for normal performance as shown in Tab. \ref{tbl:normal}.


\vspace{-0\baselineskip}
\subsection{Comparison with other methods}
\vspace{-0\baselineskip}

\begin{figure*}
\vspace{-0\baselineskip}
\centering
\includegraphics[width=\textwidth]{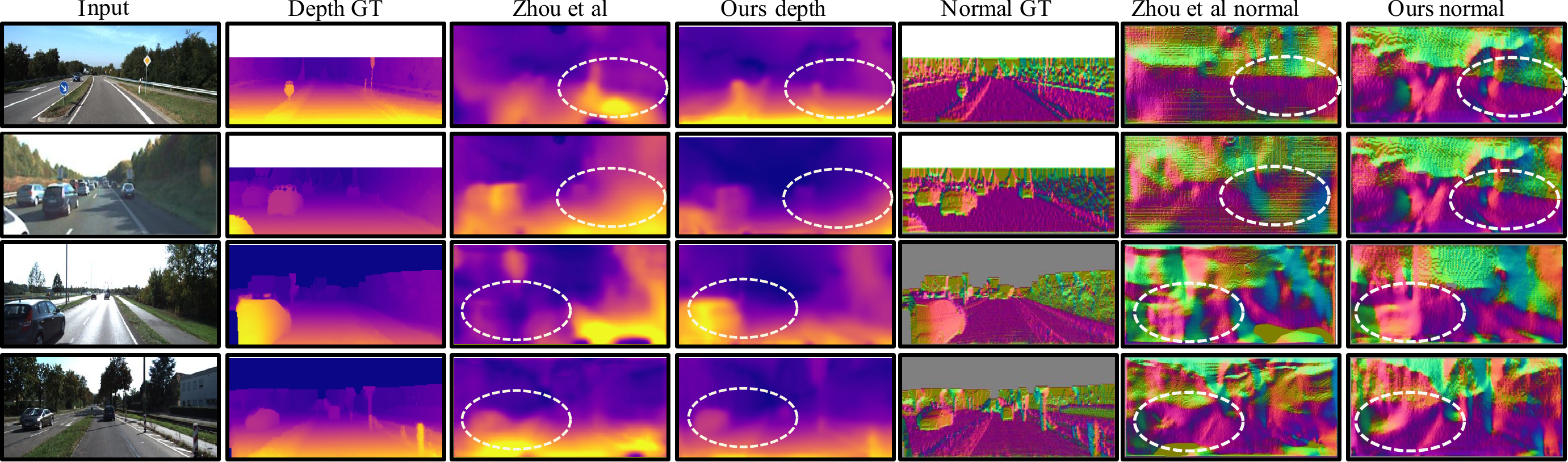}
\caption{Visual comparison between \protect\cite{zhou2017unsupervised} and ours. We use the interpolated ground truth depths and reshape the image for better visualization. For both depths and normals, our results have less artifacts, reflect the scene layouts much better (as circled in the 1st and 2nd row) and preserve more detailed structures such as cars (as circled in the 3rd and 4th row). }
\vspace{-0\baselineskip}
\label{fig:examples}
\end{figure*}

To compare with other state-of-the-arts, we show performances on both KITTI and Eigen split. The depth evaluation results are shown in Tab. \ref{tbl:sota}. Our method outperforms some supervised methods \eg \cite{eigen2014depth}, \cite{liu2016learning} and unsupervised methods \cite{zhou2017unsupervised}, \cite{kuznietsov2017semi}, while slightly worse than \cite{godard2016unsupervised} and \cite{kuznietsov2017semi}. It is worth noting that \cite{kuznietsov2017semi} utilizes the depth ground truth and \cite{godard2016unsupervised} takes stereo image pairs as input, which implies the camera motion is known. On KITTI test split, our method outperforms \cite{godard2016unsupervised} on the ``Sq Rel'' metric. As ``Sq Rel'' penalizes large depth error, due to regularization, our results has much less outlier depths. Finally, we show some qualitative results in Fig. \ref{fig:examples}.

\begin{table*}[t]
\centering
\caption{Single view depth test results on Eigen split (upper part) and KITTI split(lower part). All methods in this table use KITTI dataset for traning and the test result is capped in the range 0-80 meters. Test result on KITTI test split of Zhou et al. 2017 is generated by using their released code to train on KITTI dataset only.}
\label{tbl:sota}
\fontsize{6.5}{7}\selectfont
\bgroup
\def\arraystretch{1.4}
\begin{tabular}{lllllllllll}
\thickhline
\multirow{2}{*}{Method}                                      & \multirow{2}{*}{Test data}                        & \multicolumn{2}{l}{Supervision} & \multicolumn{4}{l}{Lower the better} & \multicolumn{3}{l}{Higher the better}               \\ \cline{3-11} 
                                                             &                                                   & Depth          & Pose           & Abs Rel  & Sq Rel & RMSE  & RMSE log & $\delta < 1.25$ & $\delta<1.25^2$ & $\delta<1.25^3$ \\ \hline
\multicolumn{1}{l|}{Train set mean}                          & \multicolumn{1}{l|}{\multirow{8}{*}{Eigen split}} & \checkmark     &                & 0.403    & 5.530  & 8.709 & 0.403    & 0.593           & 0.776           & 0.878           \\
\multicolumn{1}{l|}{\cite{eigen2014depth} Coarse}            & \multicolumn{1}{l|}{}                             & \checkmark     &                & 0.214    & 1.605  & 6.563 & 0.292    & 0.673           & 0.884           & 0.957           \\
\multicolumn{1}{l|}{\cite{eigen2014depth} Fine}               & \multicolumn{1}{l|}{}                             & \checkmark     &                & 0.203    & 1.548  & 6.307 & 0.282    & 0.702           & 0.890           & 0.958           \\
\multicolumn{1}{l|}{\cite{kuznietsov2017semi} supervised}   & \multicolumn{1}{l|}{}                             & \checkmark     &                & 0.122    & 0.763  & 4.815 & 0.194    & 0.845           & 0.957           & 0.987           \\
\multicolumn{1}{l|}{\cite{kuznietsov2017semi} unsupervised} & \multicolumn{1}{l|}{}                             &                & \checkmark     & 0.308    & 9.367  & 8.700 & 0.367    & 0.752           & 0.904           & 0.952           \\
\multicolumn{1}{l|}{\cite{godard2016unsupervised}}                  & \multicolumn{1}{l|}{}                             &                & \checkmark     & 0.148    & 1.344  & 5.927 & 0.247    & 0.803           & 0.922           & 0.964           \\
\multicolumn{1}{l|}{\cite{zhou2017unsupervised}}                    & \multicolumn{1}{l|}{}                             &                &                & 0.208    & 1.768  & 6.856 & 0.283    & 0.678           & 0.885           & 0.957           \\
\multicolumn{1}{l|}{Ours}                                    & \multicolumn{1}{l|}{}                             &                &                & 0.182    & 1.481  & 6.501 & 0.267    & 0.725           & 0.906           & 0.963           \\ \hline
\multicolumn{1}{l|}{Train set mean}                          & \multicolumn{1}{r|}{\multirow{2}{*}{}}            & \checkmark     &                & 0.398    & 5.519  & 8.632 & 0.405    & 0.587           & 0.764           & 0.880           \\
\multicolumn{1}{l|}{\cite{godard2016unsupervised}}                  & \multicolumn{1}{r|}{}                             &                & \checkmark     & 0.124    & 1.388  & 6.125 & 0.217    & 0.841           & 0.936           & 0.975           \\
\multicolumn{1}{l|}{\cite{Vijayanarasimhan17}}        & \multicolumn{1}{l|}{KITTI split}                  &                &                & -        & -      & -     & 0.340    & -               & -               & -               \\
\multicolumn{1}{l|}{\cite{zhou2017unsupervised}}                    & \multicolumn{1}{l|}{\multirow{2}{*}{}}            &                &                & 0.216    & 2.255  & 7.422 & 0.299    & 0.686           & 0.873           & 0.951           \\
\multicolumn{1}{l|}{Ours}                                    & \multicolumn{1}{l|}{}                             &                &                & 0.1648   & 1.360  & 6.641 & 0.248    & 0.750           & 0.914           & 0.969           \\ \hline
\end{tabular}
\egroup
\vspace{-0.3\baselineskip}
\end{table*}

\begin{figure}[h]
\centering
\includegraphics[width=0.5\textwidth]{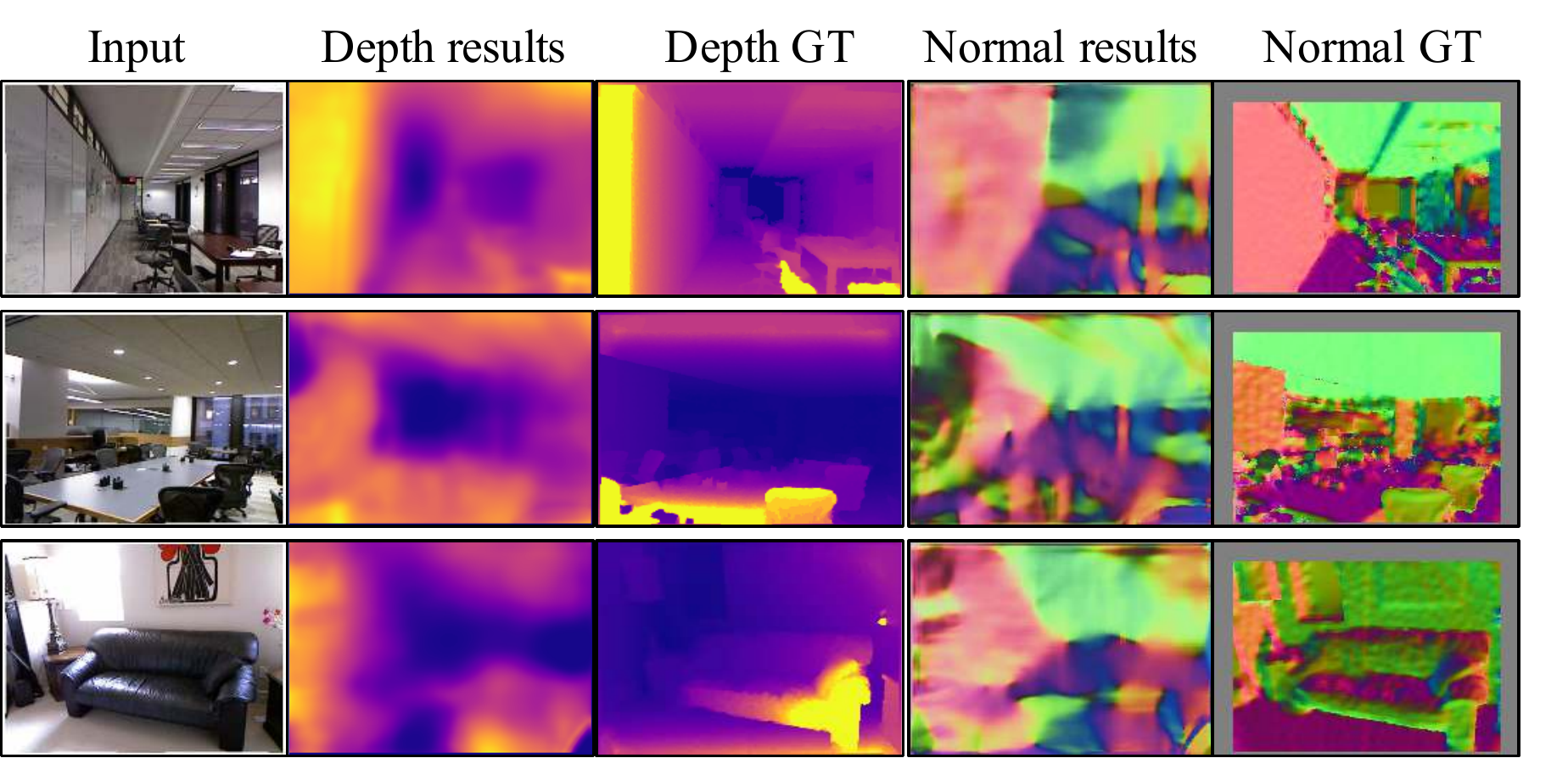}
\caption{Qualitative results of our framework on a subset of NYU v2 dataset.}
\vspace{-1.\baselineskip}
\label{fig:nyu_visual}
\end{figure}

To the best of our knowledge, there is no work reporting normal performance on the KITTI dataset. We thus compare the our normal predictions with that computed from the depth maps predicted by \cite{zhou2017unsupervised}. As shown in Tab. \ref{tbl:normal}, our method outperforms the baseline under all metrics. Additionally, to ensure the model is learned reasonably, we set up two naive baselines. ``Ground truth normal mean'' is that we set a mean normal direction for all pixels using ground truth normals. ``Pre-defined scene'' is that we separate the image to 4 parts using 4 lines connecting each image corder and image center. We set the bottom part having up-directed normal, left part having right-directed normal, right part having left-directed normal and top part with outward normals. Both of the baselines are significantly worse than our predicted model, demonstrating the correctness of the learned model.



\begin{table}[t] \small
\centering
\caption{Normal performances of our method and some baseline methods.}
\label{tbl:normal}
\fontsize{6.5}{7}\selectfont
\bgroup
\def\arraystretch{1.2}
\begin{tabular}{l|c|c|c|c|c}
\thickhline
Method                        & Mean  & Median & $11.25^{\circ}$ & $22.5^{\circ}$  & $30^{\circ}$    \\ \hline
Ground truth normal mean      & 72.39 & 64.72  & 0.031 & 0.134 & 0.243 \\
Pre-defined scene             & 63.52 & 58.93  & 0.067 & 0.196 & 0.302 \\
\cite{zhou2017unsupervised} & 50.47 & 39.16  & 0.125 & 0.303 & 0.425 \\
Ours w/o normal smoothness    & 49.30 & 36.83  & 0.138 & 0.343 & 0.436 \\
Ours                          & 47.52 & 33.98  & 0.149 & 0.369 & 0.473 \\ \hline
\end{tabular}
\egroup
\vspace{-1.0\baselineskip}
\end{table}

\vspace{-0.8\baselineskip}
\paragraph{Indoor scene exploration.}
Besides the outdoor dataset, we also try to apply our framework on the indoor NYUv2 dataset \cite{silberman2012indoor}. We use a subset for some preliminary experiments. Specifically, ``study room" is picked and split for training and testing. We first try with our baseline method~\cite{zhou2017unsupervised}, and it fails to predict any reasonable depth maps. However, as shown in Fig. \ref{fig:nyu_visual}, our framework performs reasonably good on scenes that have multiple intersecting planes. Nevertheless, we still fail on scenes that have only a clutter of object. In the future, we plan to explore more on stronger feature matching rather than just using color matching, which may facilitate the learning under cluttered scenes.

\section{Conclusion} 
In this paper, we propose an unsupervised learning framework for both depth and normal estimation via edge-aware depth-normal consistency. Our novel depth-normal regularization enforces the geoemetry consistency between different projections of the 3D scene, improving evaluation performances and also the training speed.
We present ablation experiments exploring each component of our framework and also on different scenes of images. Our results are even better than some supervised methods, and achieve state-of-the-art performance when only using monocular videos.

In the future, we would like to stick the consistency in supervised learning of both normal and depth predictions for cross supervision, which hopefully could also provides additional improvements.

\small
\bibliographystyle{aaai}
\bibliography{reference}

\begin{thebibliography}{}

\bibitem[\protect\citeauthoryear{Abadi \bgroup et al\mbox.\egroup
  }{2016}]{abadi2016tensorflow}
Abadi, M.; Agarwal, A.; Barham, P.; Brevdo, E.; Chen, Z.; Citro, C.; Corrado,
  G.~S.; Davis, A.; Dean, J.; Devin, M.; et~al.
\newblock 2016.
\newblock Tensorflow: Large-scale machine learning on heterogeneous distributed
  systems.
\newblock {\em arXiv preprint arXiv:1603.04467}.

\bibitem[\protect\citeauthoryear{Eigen and Fergus}{2015}]{eigen2015predicting}
Eigen, D., and Fergus, R.
\newblock 2015.
\newblock Predicting depth, surface normals and semantic labels with a common
  multi-scale convolutional architecture.
\newblock In {\em ICCV}.

\bibitem[\protect\citeauthoryear{Eigen, Puhrsch, and
  Fergus}{2014}]{eigen2014depth}
Eigen, D.; Puhrsch, C.; and Fergus, R.
\newblock 2014.
\newblock Depth map prediction from a single image using a multi-scale deep
  network.
\newblock In {\em NIPS}.

\bibitem[\protect\citeauthoryear{Garg, G, and Reid}{2016}]{GargBR16}
Garg, R.; G, V. K.~B.; and Reid, I.~D.
\newblock 2016.
\newblock Unsupervised {CNN} for single view depth estimation: Geometry to the
  rescue.
\newblock {\em ECCV}.

\bibitem[\protect\citeauthoryear{Geiger, Lenz, and
  Urtasun}{2012}]{geiger2012we}
Geiger, A.; Lenz, P.; and Urtasun, R.
\newblock 2012.
\newblock Are we ready for autonomous driving? the kitti vision benchmark
  suite.
\newblock In {\em CVPR}.

\bibitem[\protect\citeauthoryear{Godard, Mac~Aodha, and
  Brostow}{2017}]{godard2016unsupervised}
Godard, C.; Mac~Aodha, O.; and Brostow, G.~J.
\newblock 2017.
\newblock Unsupervised monocular depth estimation with left-right consistency.

\bibitem[\protect\citeauthoryear{Hoiem, Efros, and Hebert}{2007}]{HoiemEH07}
Hoiem, D.; Efros, A.~A.; and Hebert, M.
\newblock 2007.
\newblock Recovering surface layout from an image.
\newblock In {\em ICCV}.

\bibitem[\protect\citeauthoryear{Ioffe and Szegedy}{2015}]{ioffe2015batch}
Ioffe, S., and Szegedy, C.
\newblock 2015.
\newblock Batch normalization: Accelerating deep network training by reducing
  internal covariate shift.
\newblock In {\em ICML}.

\bibitem[\protect\citeauthoryear{Jaderberg \bgroup et al\mbox.\egroup
  }{2015}]{jaderberg2015spatial}
Jaderberg, M.; Simonyan, K.; Zisserman, A.; et~al.
\newblock 2015.
\newblock Spatial transformer networks.
\newblock In {\em Advances in Neural Information Processing Systems},
  2017--2025.

\bibitem[\protect\citeauthoryear{Jia}{2006}]{jia2006using}
Jia, Z.
\newblock 2006.
\newblock Using cross-product matrices to compute the svd.
\newblock {\em Numerical Algorithms} 42(1):31--61.

\bibitem[\protect\citeauthoryear{Karsch, Liu, and Kang}{2014}]{karsch2014depth}
Karsch, K.; Liu, C.; and Kang, S.~B.
\newblock 2014.
\newblock Depth transfer: Depth extraction from video using non-parametric
  sampling.
\newblock {\em IEEE transactions on pattern analysis and machine intelligence}
  36(11):2144--2158.

\bibitem[\protect\citeauthoryear{Kong and Black}{2015}]{kong2015intrinsic}
Kong, N., and Black, M.~J.
\newblock 2015.
\newblock Intrinsic depth: Improving depth transfer with intrinsic images.
\newblock In {\em ICCV}.

\bibitem[\protect\citeauthoryear{Kuznietsov, Stuckler, and
  Leibe}{2017}]{kuznietsov2017semi}
Kuznietsov, Y.; Stuckler, J.; and Leibe, B.
\newblock 2017.
\newblock Semi-supervised deep learning for monocular depth map prediction.

\bibitem[\protect\citeauthoryear{L.~Ladicky, Pollefeys, and
  others}{2014}]{zeisl2014discriminatively}
L.~Ladicky, Zeisl, B.; Pollefeys, M.; et~al.
\newblock 2014.
\newblock Discriminatively trained dense surface normal estimation.
\newblock In {\em ECCV}.

\bibitem[\protect\citeauthoryear{Ladicky, Shi, and
  Pollefeys}{2014}]{ladicky2014pulling}
Ladicky, L.; Shi, J.; and Pollefeys, M.
\newblock 2014.
\newblock Pulling things out of perspective.
\newblock In {\em CVPR}.

\bibitem[\protect\citeauthoryear{Laina \bgroup et al\mbox.\egroup
  }{2016}]{laina2016deeper}
Laina, I.; Rupprecht, C.; Belagiannis, V.; Tombari, F.; and Navab, N.
\newblock 2016.
\newblock Deeper depth prediction with fully convolutional residual networks.
\newblock In {\em 3D Vision (3DV), 2016 Fourth International Conference on},
  239--248.
\newblock IEEE.

\bibitem[\protect\citeauthoryear{Li \bgroup et al\mbox.\egroup
  }{2015}]{li2015depth}
Li, B.; Shen, C.; Dai, Y.; van~den Hengel, A.; and He, M.
\newblock 2015.
\newblock Depth and surface normal estimation from monocular images using
  regression on deep features and hierarchical crfs.
\newblock In {\em CVPR}.

\bibitem[\protect\citeauthoryear{Li}{2017}]{li2017pyramidal}
Li, Y.
\newblock 2017.
\newblock Pyramidal gradient matching for optical flow estimation.
\newblock {\em arXiv preprint arXiv:1704.03217}.

\bibitem[\protect\citeauthoryear{Liu \bgroup et al\mbox.\egroup
  }{2016}]{liu2016learning}
Liu, F.; Shen, C.; Lin, G.; and Reid, I.
\newblock 2016.
\newblock Learning depth from single monocular images using deep convolutional
  neural fields.
\newblock {\em IEEE transactions on pattern analysis and machine intelligence}
  38(10):2024--2039.

\bibitem[\protect\citeauthoryear{Liu, Shen, and Lin}{2015}]{Liu_2015_CVPR}
Liu, F.; Shen, C.; and Lin, G.
\newblock 2015.
\newblock Deep convolutional neural fields for depth estimation from a single
  image.
\newblock In {\em CVPR}.

\bibitem[\protect\citeauthoryear{Long, Shelhamer, and
  Darrell}{2015}]{long2015fully}
Long, J.; Shelhamer, E.; and Darrell, T.
\newblock 2015.
\newblock Fully convolutional networks for semantic segmentation.
\newblock In {\em CVPR}.

\bibitem[\protect\citeauthoryear{Mayer \bgroup et al\mbox.\egroup
  }{2016}]{mayer2016large}
Mayer, N.; Ilg, E.; Hausser, P.; Fischer, P.; Cremers, D.; Dosovitskiy, A.; and
  Brox, T.
\newblock 2016.
\newblock A large dataset to train convolutional networks for disparity,
  optical flow, and scene flow estimation.
\newblock In {\em CVPR}.

\bibitem[\protect\citeauthoryear{Mur-Artal, Montiel, and
  Tardos}{2015}]{mur2015orb}
Mur-Artal, R.; Montiel, J. M.~M.; and Tardos, J.~D.
\newblock 2015.
\newblock Orb-slam: a versatile and accurate monocular slam system.
\newblock {\em IEEE Transactions on Robotics} 31(5):1147--1163.

\bibitem[\protect\citeauthoryear{Newcombe, Lovegrove, and
  Davison}{2011}]{NewcombeLD11}
Newcombe, R.~A.; Lovegrove, S.; and Davison, A.~J.
\newblock 2011.
\newblock {DTAM:} dense tracking and mapping in real-time.
\newblock In {\em ICCV}.

\bibitem[\protect\citeauthoryear{Prados and Faugeras}{2006}]{prados2006shape}
Prados, E., and Faugeras, O.
\newblock 2006.
\newblock Shape from shading.
\newblock {\em Handbook of mathematical models in computer vision}  375--388.

\bibitem[\protect\citeauthoryear{Schwing \bgroup et al\mbox.\egroup
  }{2013}]{DBLP:conf/iccv/SchwingFPU13}
Schwing, A.~G.; Fidler, S.; Pollefeys, M.; and Urtasun, R.
\newblock 2013.
\newblock Box in the box: Joint 3d layout and object reasoning from single
  images.
\newblock In {\em ICCV}.

\bibitem[\protect\citeauthoryear{Silberman \bgroup et al\mbox.\egroup
  }{2012}]{silberman2012indoor}
Silberman, N.; Hoiem, D.; Kohli, P.; and Fergus, R.
\newblock 2012.
\newblock Indoor segmentation and support inference from rgbd images.
\newblock {\em ECCV}.

\bibitem[\protect\citeauthoryear{Srajer \bgroup et al\mbox.\egroup
  }{2014}]{DBLP:conf/3dim/SrajerSPP14}
Srajer, F.; Schwing, A.~G.; Pollefeys, M.; and Pajdla, T.
\newblock 2014.
\newblock Match box: Indoor image matching via box-like scene estimation.
\newblock In {\em 3DV}.

\bibitem[\protect\citeauthoryear{Vijayanarasimhan \bgroup et al\mbox.\egroup
  }{2017}]{Vijayanarasimhan17}
Vijayanarasimhan, S.; Ricco, S.; Schmid, C.; Sukthankar, R.; and Fragkiadaki,
  K.
\newblock 2017.
\newblock Sfm-net: Learning of structure and motion from video.
\newblock {\em CoRR} abs/1704.07804.

\bibitem[\protect\citeauthoryear{Wang \bgroup et al\mbox.\egroup
  }{2015}]{DBLP:conf/cvpr/WangSLCPY15}
Wang, P.; Shen, X.; Lin, Z.; Cohen, S.; Price, B.~L.; and Yuille, A.~L.
\newblock 2015.
\newblock Towards unified depth and semantic prediction from a single image.
\newblock In {\em CVPR}.

\bibitem[\protect\citeauthoryear{Wang \bgroup et al\mbox.\egroup
  }{2016}]{peng2016depth}
Wang, P.; Shen, X.; Russell, B.; Cohen, S.; Price, B.~L.; and Yuille, A.~L.
\newblock 2016.
\newblock {SURGE:} surface regularized geometry estimation from a single image.
\newblock In {\em NIPS}.

\bibitem[\protect\citeauthoryear{Wang, Fouhey, and
  Gupta}{2015}]{wang2015designing}
Wang, X.; Fouhey, D.; and Gupta, A.
\newblock 2015.
\newblock Designing deep networks for surface normal estimation.
\newblock In {\em CVPR}.

\bibitem[\protect\citeauthoryear{Wu and others}{2011}]{wu2011visualsfm}
Wu, C., et~al.
\newblock 2011.
\newblock Visualsfm: A visual structure from motion system.

\bibitem[\protect\citeauthoryear{Xiao, Owens, and
  Torralba}{2013}]{xiao2013sun3d}
Xiao, J.; Owens, A.; and Torralba, A.
\newblock 2013.
\newblock Sun3d: A database of big spaces reconstructed using sfm and object
  labels.
\newblock In {\em ICCV}.

\bibitem[\protect\citeauthoryear{Xie, Girshick, and
  Farhadi}{2016}]{xie2016deep3d}
Xie, J.; Girshick, R.; and Farhadi, A.
\newblock 2016.
\newblock Deep3d: Fully automatic 2d-to-3d video conversion with deep
  convolutional neural networks.
\newblock In {\em ECCV}.

\bibitem[\protect\citeauthoryear{Zhou \bgroup et al\mbox.\egroup
  }{2017}]{zhou2017unsupervised}
Zhou, T.; Brown, M.; Snavely, N.; and Lowe, D.~G.
\newblock 2017.
\newblock Unsupervised learning of depth and ego-motion from video.
\newblock In {\em CVPR}.

\end{thebibliography}

\end{document}